\documentclass[journal,twoside,web]{ieeecolor}
\usepackage{tmi}
\usepackage{cite}
\usepackage{amsmath,amssymb,amsfonts}
\usepackage{algorithmic}
\usepackage{graphicx}
\usepackage{textcomp}

\usepackage{graphicx}
\usepackage{amsmath}
\usepackage{amssymb}
\usepackage{booktabs}
\usepackage{caption}
\usepackage{subcaption}

\usepackage[pagebackref,breaklinks,colorlinks,bookmarks=false]{hyperref}
\newcommand{\norm}[1]{\left\lVert#1\right\rVert}

\begin{document}

\title{Graph-in-Graph (GiG): Learning interpretable latent graphs in non-Euclidean domain for biological and healthcare applications}

\author{
Kamilia Mullakaeva, Luca Cosmo, Anees Kazi, Seyed-Ahmad Ahmadi, Nassir Navab, Michael M. Bronstein
\thanks{K. Mullakaeva is with the Department of Computer Science, Technical University of Munich, Munich, Germany (e-mail: kamilia.mullakaeva@tum.de). }
\thanks{A. Kazi is with the Department of Computer Science, Technical University of Munich, Munich, Germany (e-mail: anees.kazi@tum.de). }
\thanks{L. Cosmo is with the Department of Environmental Sciences, Informatics and Statistics,
Ca' Foscari University of Venice, Venice, Italy. He is also collaborating with the Informatics Departement, USI University of Lugano, Lugano, Switzerland (e-mail: luca.cosmo@unive.it).}
\thanks{S.-A. Ahmadi is with NVIDIA, Munich, Germany. (e-mail: ahmadi@cs.tum.edu).}
\thanks{N. Navab is with the Department of Computer Science, Technical University of Munich, Munich, Germany and also with the Whiting School of Engineering, Johns Hopkins University, Baltimore, USA (e-mail: nassir.navab@tum.de).}
\thanks{M. M. Bronstein is with the University of Oxford (e-mail: michael.bronstein@cs.ox.ac.uk)}}

\maketitle
\begin{abstract}
Graphs are a powerful tool for representing and analyzing unstructured, non-Euclidean data ubiquitous in the healthcare domain. Two prominent examples are molecule property prediction and brain connectome analysis. Importantly, recent works have shown that considering relationships between input data samples have a positive regularizing effect for the downstream task in healthcare applications. These relationships are naturally modeled by a (possibly unknown) graph structure between input samples. In this work, we propose Graph-in-Graph (GiG), a neural network architecture for protein classification and brain imaging applications that exploits the graph representation of the input data samples and their latent relation. We assume an initially unknown latent-graph structure between graph-valued input data and propose to learn end-to-end a parametric model for message passing within and across input graph samples, along with the latent structure connecting the input graphs.
Further, we introduce a degree distribution loss that helps regularize the predicted latent relationships structure. This regularization can significantly improve the downstream task. Moreover, the obtained latent graph can represent patient population models or networks of molecule clusters, providing a level of interpretability and knowledge discovery in the input domain of particular value in healthcare.
\end{abstract}

\begin{IEEEkeywords}
Graph Deep Learning, Knowledge Discovery
\end{IEEEkeywords}

\section{Introduction}
\label{sec:introduction}
\IEEEPARstart{R}{ecently}, applications with graph-based data have rapidly increased in many domains such as computer vision \cite{methods_app_review}, Computer graphics \cite{wang2019dynamic}, Physics \cite{shlomi2020graph}, Chemistry \cite{fung2021benchmarking} and now healthcare \cite{healthcare_management}. 
Graph Neural Networks (GNNs) have proven to be a very powerful tool to process non-Euclidean unstructured data \cite{overview_GNN}, especially in healthcare applications such disease prediction \cite{kazi2019graph}, drug interaction \cite{lin2020kgnn} and discovery \cite{li2017learning}, brain connectome analysis \cite{kim2021learning}, multi-modal data analysis \cite{Cosmo_2020}, just to name a few.

Most of the GNN models for graph classification consider each input graph individually. They aggregate information between neighboring nodes via message passing, to obtain new node or graph representations to be used for the final classification layers. On the other hand, recent literature on disease prediction testify that considering information coming from similar patients, in form of a population level graph, is crucial to enhance performance of the model \cite{kazi2017automatic}.
This is the case, for instance, of the method proposed in \cite{parisot2018disease} for brain connectome analysis, where single patient's brain graphs are converted into a vector representation and later leveraged in a population level graph with each node as an individual patient associated to a feature vector representing the individual brain graph.

Despite showing the importance of considering the population level information to the downstream task, all these methods operates on a flattened representation of the input data, obtained by some preprocessing step to embed the input graph on a vectorial space. As a consequence, the structural information from the individual graphs is often lost as the method only focuses on the resulting feature vector. Moreover, the input objects in healthcare or molecular prediction incorporate crucial structural and functional information within the graph structure itself that could be beneficial to the task at hand.   
In this paper, we propose Graph-in-Graph (GiG), a graph learning architecture which exploits both structural information and node level features of the input graph samples, together with leveraging population level information by learning a graph between input samples.
Importantly, learning the population level graph provides a form of knowledge discovery through the learned latent graph, and enables reasoning about made decisions through learned neighborhoods around input samples.
Note that, most deep learning works optimize the downstream task only, and analyze the latent structure post-hoc, through mapping techniques like t-SNE \cite{vandermaaten2008tsne} or UMAP \cite{mcinnes2020umap}. Instead, we aim to learn the downstream task and construct the latent graph structure end-to-end. To the best of our knowledge, the combination of latent graph learning and graph-valued input data has not been investigated so far.

We also propose a degree distribution loss that regularizes the predicted latent relationship structure. This loss provides a parameter that allows the user to influence the connectivity degrees across samples. This can lead to better classification performance, but more importantly, it can facilitate knowledge discovery by forcing the model to limit the neighborhood sizes around input samples. 

We evaluate our method on real-world datasets from different domains, one medical and two bioinformatic, and provide a comparison with state-of-the-art graph classification models.

\section{Related Work}
\label{sec:related_work}

Protein classification, brain connectome classification, and toxicology analysis are among the most relevant biomedical applications in which the input data is naturally represented by a graph structure.

In the domain of protein classification, one of the first learning approaches was based on extracting feature vectors from proteins and classifying them into enzymes or non-enzymes using support vector machines (SVM) \cite{DOBSON2003771}. Later, variations were proposed, e.g. using C-Support Vector Machines \cite{cortes1995support} with graph kernels \cite{10.1093/bioinformatics/bti1007}. Recently, the use of Graph Convoutional Neural Networks (GCNNs) has been proposed as a general framework for graph and node representation learning, including for proteins classification {\cite{wang2019dynamic,xu2019powerful, DBLP:journals/corr/abs-1909-10086}}. GNNs have shown to be able to capture local structures which are usually characteristic of the graph representation of these data \cite{DBLP:conf/nips/NikolentzosV20,bouritsas2021partition,DBLP:journals/corr/abs-2112-07436}.

In particular, Zhang et al. \cite{zhang2019hierarchical} reached SOTA results with the use of pooling operators and structure learning, which down-sample graph data and obtain a better embedding for the downstream classification tasks . 

Brain connectome analysis requires pre-processing of brain signal time-series from image/sensor data, in order to extract a connectivity graph between brain regions. In a recent work \cite{PERVAIZ2020116604}, the application of different models such as GCN, BrainNetCNN, and ElasticNet were compared on brain connectivity graphs from the Human Connectome Project (HCP) dataset \cite{VANESSEN201362}, to suggest a unified brain data processing algorithm.

Toxicity prediction for molecules is a challenging task, and is well-studied thanks to open-access datasets like the Tox21 Data Challenge\footnote{Tox21 Data Challenge URL: \url{https://tripod.nih.gov/tox21/challenge/about.jsp}}. For example, DeepTox \cite{DeepTox} suggested to use Deep Neural Networks (DNNs) and multi-task learning, while Dmlab \cite{Dmlab} and Microsomes \cite{Microsomes} provided tree-based ensemble methods as a solution. Class balance issues in the input data were addressed in \cite{balancingTox21}, by suggesting different bagging approaches through resampling techniques. These models used input data in the Simplified Molecular-Input Line-Entry (SMILES) format. However, SMILES  does not optimally preserve the molecular structure \cite{Guo_2021}. Thus, the first bridge works \cite{zaslavskiy2018toxicblend, GraSeq} combined different molecular fingerprints and molecular graphs as input. Recently a group of models were introduced that use the molecular graph as input \cite{lu2019molecular, Mansimov_2019, Guo_2021}. For example, \cite{Guo_2021} suggested meta-learning framework to obtain better results on a small number of samples. Proper pooling operations are another way to improve the graph representation and to present the entire graph structure. For example, Graph Multiset Transformer (GMT) \cite{baek2021accurate} uses a multi-head attention mechanism to better preserve graph structural information.

A common factor in all these applications is the recent proliferation of methods applying GCNNs directly on the input graph representation. GCNNs have indeed shown to be a powerfull tool to compute optimal nodes/graphs representations, showing to be the weapon of choice for learning tasks in several domains, such as social sciences \cite{zhang2018link,qi2018learning}, computer vision and graphics \cite{qi20173d,monti2017geometric,wang2019dynamic}, physical  \cite{choma2018graph,duvenaud2015convolutional,gilmer2017neural}, as well as medical and biological sciences \cite{parisot2018disease,parisot2017spectral,mellema2019multiple,kazi2019inceptiongcn,zitnik2018modeling,Zitnik19,gainza2019deciphering}. 

Another breakthrough in graph classification, especially in the domain of brain imaging and disease prediction, has been considering relationships between patients in the form of a population level graph \cite{parisot2017spectral,parisot2018disease,vivar2018multi,kazi2019graph,kazi2019self,s20216001}. 
Parisot et al. \cite{parisot2017spectral,parisot2018disease} suggested to build a population graph as a sparse graph, where nodes represent imaging data and edges phenotypic information. Later, Kazi et al. \cite{kazi2019self} proposed to construct several graphs  each corresponding to one of the demographic element and combine them in self-attention layer. Similarly, in \cite{s20216001} was proposed to combine separately constructed population graphs for diagnosing autism spectrum disorder. More recently, authors from \cite{Cosmo_2020} proposed a continuous differentiable graph module (cDGM) to learn a weighted adjacency matrix representing the population-level graph by training end-to-end the graph generation and the classification weights of the model. In \cite{kazi2020differentiable}, cDGM is expanded by training a probabilistic graph generator and sampler able to handle sparse population graphs. However, all these methods assume vector-valued input data, either in form of a flattened feature vector at input, or in form of an embedding vector obtained from an input module. 

However, all the methods exploiting a population level graph assume vector-valued input data of the input samples, either in form of a flattened feature vector at input or in form of an embedding vector obtained from an input module, even when the input data are naturally represented by a graph structure.
On the other hand, SOTA methods for graph classification consider input graphs in isolation and none of the methods working in molecule classification consider a latent graph between molecules to obtain better representations. We argue that preserving the initial graph structures at input, and simultaneously learning their embedding in an end-to-end fashion, can be beneficial to obtain a better representation for downstream tasks, and to extract only task-relevant information from feature vectors. Moreover, most datasets in the healthcare or bioinformatics domain suffer from an insufficient amount of data and missing information. In these cases, being able to recover missing information from neighboring nodes, which is enabled through the learned latent graph, is a crucial factor to provide a correct classification.

This paper aims exactly at filling this gap, proposing a method that works directly on the graph representation of the input data and learns a population level latent graph in a end-to-end fashion by optimizing the classification task loss.

\section{Method}
\begin{figure*}
  \centering
    \includegraphics[width=1\textwidth]{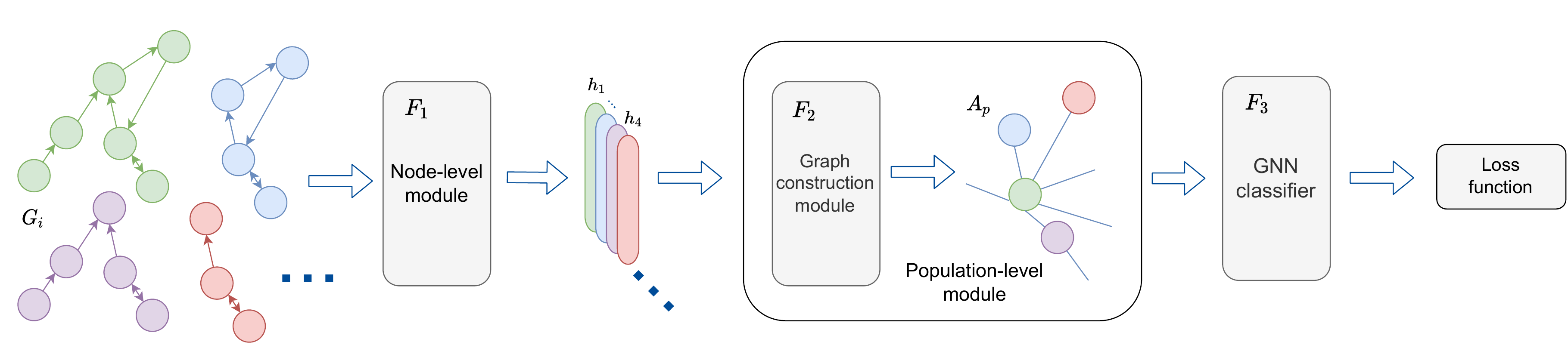}
    \caption{Graph-in-Graph architecture consists of three parts: $F_1$ represents the node-level module, $F_2$ - the population-level module, and $F_3$ - GNN classifier. As input model has graphs $\mathbf{G_i}$, then its representation $\mathbf{h_i}$ is obtained from $F_1$.}
    \label{fig:Graph_in_Graph_architecture}
\end{figure*}
The input to our model is a set of $N$ graphs, $\mathbf{G} = {\mathbf{G_1}, \mathbf{G_2}, ..., \mathbf{G_N}}$. Each $i^{th}$ graph is defined as $\mathbf{G_i}=(\mathbf{V_i}, \mathbf{E_i}, \mathbf{X_i})$, 
where $\mathbf{V_i}$ and $\mathbf{E_i}$ are vertices and edges of the graph and $\mathbf{X_i}\in \mathbb{R}^{\|\mathbf{V_i}\| \times D}$ is the node features matrix with $D$ being number of features. 
The output of our model is a class probability vector  $\mathbf{p}\in \mathbb{R}^{C}$ for each input graph, where $C$ is the numbers of possible classes to which the input graph can be categorized. 

\subsection{Graph-in-Graph model}
In this section we provide the technical details of the proposed GiG model (Fig. \ref{fig:Graph_in_Graph_architecture}).
The proposed model consists of three parts: 1) the node-level module $F_1$; 2) the population-level module $F_2$; 3) the final GNN classifier $F_3$ (Fig. \ref{fig:Graph_in_Graph_architecture}). In a nutshell $F_1$ computes input graph representations, $F_2$  focuses on learning the latent connections between the input graphs, and $F_3$ takes as input both the graph representations from $F_1$ and the latent structure from $F_2$ and provides the predictions of the downstream task as output. In the following subsections we provide technical details of each of the three parts.

\subsubsection{Node-level module $F_1$}
The node-level module is designed to directly handle non-Euclidian data (i.e. graphs) as input and capture only task-relevant features and structural information during the training. Mathematically $F_1$ is defined as, $ \mathbf{h_i}= F_{1}(\mathbf{G_i})$, where $\mathbf{h} = [\mathbf{h_1},\dots,\mathbf{h_N}]$ are output graph representations in a H dimensional latent space, with $\mathbf{h_i} \in \mathbb{R}^{1\times H}$. $F_1$ is implemented as a graph convolution based neural network composed by a set of graph convolutional layers followed by a pooling operator over all node features.
The specific choice of the graph convolutional operator to use in $F_1$ depends on the dataset and the task at hand.
 
\subsubsection{Population-level module $F_2$} 
The population-level module targets learning the latent population graph. The output of $F_1$ i.e $\mathbf{h} = [\mathbf{h_0},...\mathbf{h_i},...,\mathbf{h_N}]$ is fed to $F_2$ as input.
Each $i^{th}$ node in the population level graph corresponds to the $i^{th}$ input graph of the previous module, and it is represented by its representation $\mathbf{h_i}$. Formally, $F_2$ is a function defined as $\mathbf{A_p} = F_2(\mathbf{h})$, where $\mathbf{A_p} \in \left (0,1  \right )^{N\times N}$ is the weighted adjacency matrix encoding the population-level graph. 
Inspired by the population graph learning strategy proposed in \cite{Cosmo_2020}, we develop our LGL technique to let our model learn the population-level graph in a end-to-end fashion together with minimizing the downstream task loss.

We learn the population-level graph by embedding each input graph representation $\mathbf{h_i}$ in a lower dimensional latent space in which the closeness of of two node features indicates the existence of an edge between the two nodes in the population graph.

To project the input graphs to this latent space we define a learnable function $g$ which takes as input the original graph representations $\mathbf{h_i}$ and outputs a new representation $\Tilde{\mathbf{h_i}}$ belonging to this new latent space, $\Tilde{\mathbf{h_i}} = g(\mathbf{h_i})$. In our case, we use an MLP as our $g$ function. The output of this module is then a weighted adjacency matrix $\mathbf{A_p}$ built from input graph representations $\Tilde{\mathbf{h}}$. Precisely, the edge $a_{ij}$ between the nodes $i$ and $j$ in $\mathbf{A_p}$ is defined as:
\begin{equation}\label{eq:adj}
a_{ij} =\frac{1}{1+e^{-t \norm{\Tilde{\mathbf{h_i}}-\Tilde{\mathbf{h_j}}}_2+\theta}} \; ,
\end{equation}
where $\theta$ and $t$ are learnable soft-threshold and temperature parameters. 

\subsubsection{GNN classifier $F_3$}
The output of the population-level module $\mathbf{A_p}$ is exploited by the final module (i.e. the GNN classifier) to provide the final prediction according to the downstream  task. This later module is composed by a function $\mathbf{p}=F_3(\mathbf{h},\mathbf{A_p})$, with $\mathbf{p}=[\mathbf{p_1},\dots,\mathbf{p_N}]$, that takes as input the individual graph representations $\mathbf{h}$ and the learned population-graph $\mathbf{A_p}$ and outputs a class probability vector $\mathbf{p_i}$ for each of the input graphs. Specifically, $F_3$ is composed of a set of GNN layers followed by node pooling and a fully-connected layer to obtain the output probabilities. $F_3$ being a graph convolution based module is capable of leveraging input sample similarities and obtain better representations for the downstream task.

\begin{figure}[b]
\begin{subfigure}{.240\textwidth}
  \centering
  \includegraphics[height=25mm]{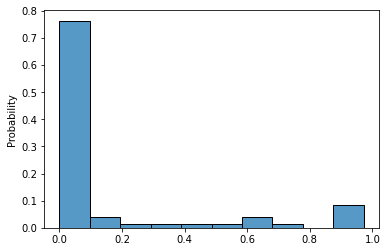}
  \caption{LGL}
  \label{fig:proteins_weights_degree_distr_a}
\end{subfigure}
\begin{subfigure}{.240\textwidth}
  \centering
  \includegraphics[height=25mm]{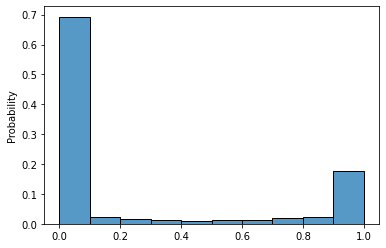}
  \caption{LGL+KL}
  \label{fig:proteins_weights_degree_distr_b}
\end{subfigure}
\caption{Distribution of adjacency matrix $\mathbf{A_p}$ values for PROTEINS$_{29}$ dataset with (b) and without (a) the degree distribution loss $KL_{loss}$. The x-axis indicates values of $\mathbf{A_p}$, y-axis shows the values occurrence probability.}
\label{fig:proteins_weights_degree_distr}
\end{figure}

\subsection{Degree distribution loss}
\label{loss_definition}
In this section we propose a customised loss capable of not only achieving the down stream task but also ensuring the quality of population graph. We experimentally observed that optimizing the population-level module just using the task classification loss tends to produce sparser graphs. This behavior can be seen from the left bar plot in Fig. \ref{fig:proteins_weights_degree_distr} showing the values distribution of $\mathbf{A_p}$ optimizing with the plain cross-entropy loss. As a consequence, the resulting graph is composed by many disconnected components and isolated nodes (Fig. \ref{fig:proteins_nodes_degree_distr}, \ref{fig:hcp_names_population_graph_comparison}) which makes it difficult to reason about the predicted population graph structure.

To mitigate this problem we propose to add a Degree Distribution Loss as a regularizer which helps produce more meaningful population graphs. 
The population-level adjacency matrix $\mathbf{A_p}$ is regularised by adding a prior to the desired node degrees distribution. The desired degree distribution is promoted by an additional Kullback–Leibler divergence loss (KL loss) between the computed degree distribution over $\mathbf{A_p}$ and the target distribution.  
In this work, we choose a Gaussian distribution as a target distribution, centered at the desired average node degree. This enforces a sparse graph, while leaving the possibility of a few nodes having dense connections. We let the parameters (mean, standard deviation) of the Gaussian target distribution to be optimized during the training, in order to give the model more flexibility in adapting the population graph for optimal mean and standard deviation of node degrees. 

On a symmetric adjacency matrix $\mathbf{A}$, the degree of the $i^{th}$ node can be computed as the sum of the entries of the $i^{th}$ row. Since $\mathbf{A_p}$ represents a weighted fully-connected graph, the actual degree of each node is always $N-1$. To overcome this limitation, we exploit the fact that \eqref{eq:adj} has the effect to push values of $\mathbf{A_p}$ either toward 0 or 1 and consider as actual edges of the graph only the edges with a weight greater that 0.5:

\begin{equation}
\mathbf{\bar{A}} = \mathbf{A_p}*(\mathbf{A_p}>0.5)
\label{eq_ A_Ap}
\end{equation}
where $\mathbf{\bar{A}} \in \mathbb{R}^{N\times N}$ is the adjacency matrix used to compute the node degree of each node as 
\begin{equation}
\bar d_{j}= \sum_{i=1}^N \mathbf{\bar A}_{i,j}
\label{eq:sum_adj}
\end{equation}
with  $ \bar d_j \in \mathbb{R}^{N} $. Since $\bar d_j$ can assume a continuous value between $0$ and $N-1$, to compute the discrete degree distribution we perform a soft assignment between $d_j$ and the discrete node degrees value $c_i\in[0,1,\dots,N-1]$ as:
\begin{equation}
    S_{i, j} = \frac{e^{\frac{(c_{i} -\bar{d_{j}})^2}{\sigma^2}}}{\sum_k e^{\frac{(c_{k} -\bar{d_{j}})^2}{\sigma^2}} }
\end{equation}

where $\sigma$ is a hyperparameter that we experimentally set to 0.6. The value corresponding to the degree $c_i$ in the computed node degree distribution is:
\begin{equation}
p_{i} = \frac{ \sum_{j} S_{i, j} }{\sum_{k,j} S_{k, j} }
\end{equation}

Finally, the $ KL_{loss}$ term is defined as
\begin{equation}
KL_{loss} = D_{KL}(p,q) 
\end{equation}
where $q$ is the target normal discrete distribution with learnable parameters. The KL-loss is thus added to the Cross-Entropy loss used to train the classifier as a penalty term weighted by $\alpha$:
\begin{equation}
loss = CE_{loss} + \alpha KL_{loss}
\label{eq:loss_sum}
\end{equation}
\begin{table*}[th!]
\centering
\caption[Comparison of results for dataset]{Comparison of dataset with several baselines and SOTA methods. ElasticNet \cite{ElasticNet}, HGP-SL \cite{zhang2019hierarchical}, ToxicBlend \cite{zaslavskiy2018toxicblend} SOTA methods for HCP, PROTEINS and Tox21 respectively. The top three  performance is shown in bold face, blue and red.}
\begin{tabular}{@{}l|cc|ccc|cc@{}}
&\multicolumn{2}{c}{\textit{Without population} graph} &\multicolumn{3}{|c}{\textit{Non-learned population graph}} &\multicolumn{2}{|c}{\textit{Learned population graph}}\\
\hline
&SOTA &GCN &Random &KNN &GiG DGCNN &GiG LGL &GiG LGL+KL \\\hline
HCP (acc \%) &85.5 ± NA &84.2 ± 2.5 &43.7 ± 8.4 &39.6 ± 0.0 &\textcolor{red}{89.0 ± 3.3} &\textcolor{blue}{89.4 ± 1.3} &\textbf{89.7 ± 1.3} \\
PROTEINS$_{29}$ (acc \%) &\textbf{84.9 ± 1.6} &80.0 ± 2.5 &53.7 ± 8.4 &74.9 ± 2.7 &72.4 ± 2.2 &\textcolor{red}{82.0 ± 1.6} &\textcolor{blue}{84.8 ± 1.1} \\
Tox21 (acc \%) &80.7 ± NA &75.0 ± 0.8 &80.1 ± 3.9 &80.1 ± 2.0 &\textbf{85.1 ± 2.1} &\textcolor{blue}{84.8 ± 1.7} &\textcolor{red}{82.3 ± 2.3} \\
\hline
\end{tabular}
\label{tab:acc_tab}
\end{table*}

\section{Experiments and results}
In this section, we show experiments on biological, medical and chemical applications where each input sample is represented as a graph.
In particular we choose PROTEINS \cite{KKMMN2016}, HCP \cite{VANESSEN201362} and Tox21 \cite{wu2018moleculenet}. Further, we make use of the publicly available benchmark datasets D\&D \cite{DOBSON2003771}, NCI1 \cite{nci1} and ENZYMES \cite{enzymes} to compare with other general purpose graph classification methods. These datasets are commonly used in the evaluation of graph classification models for binary and multi-class classification. In all the benchmark datasets we follow the training/testing procedure described in \cite{errica2020fair} for a fair comparision.

We show two variants of the GiG model. In the first variant (LGL) the population-level graph is learned end-to-end using just the cross-entropy loss for the final classification task as in \cite{Cosmo_2020}. In the second variant (LGL+KL) the model incorporates the proposed node degree distribution loss \eqref{loss_definition} along with learning the population-level graph. 
In order to better validate the benefit of the population level module, we also investigate different possibilities to build or learn the population graph.

\subsection{Datasets}

In this section, we give the details of the datasets used for our experimental evaluation.

\textbf{HCP.\hspace{1mm}} \label{sec:hcp-datasets}
The Human Connectome Project (HCP) is a medical dataset consisting of 1003 complete resting-state functional MRI (fMRI) runs released in 2017 \cite{VANESSEN201362}. The task is to predict the gender of each patient based on the corresponding brain fMRI data.
The brain graphs are provided within the dataset as the correlation matrix between fMRI signals from each parcels. Each brain consist of 200 parcels, and hence comes with an affinity matrix of size $200 \times 200$.

\textbf{PROTEINS.\hspace{1mm}}\label{sec:proteins-datasets}
PROTEINS is a binary classification dataset \cite{10.1093/bioinformatics/bti1007} implemented by the TU Dortmund University \cite{KKMMN2016}. Here, the task is to classify the protein as an enzyme or non-enzyme.
We used two versions of the PROTEINS dataset: PROTEINS$_{29}$ and PROTEINS$_3$, where the index indicates the number of features used in the feature matrix. 
For population graph analyses, we use the CATH Superfamily information \cite{ORENGO19971093, CATH2} for each protein, which was retrieved from \cite{10.1093/nar/gkaa1079}. Importantly, CATH labels were not used during the training stage. Instead, we used CATH labels only to investigate and discuss potential knowledge discovery of population-level graphs.
The CATH hierarchy consists of several hierarchical class levels. We consider only the first class level (C-level), consisting of three major classes: "mainly-alpha", "mainly-beta", and "alpha-beta". 

\textbf{Tox21.\hspace{1mm}}\label{sec:tox21-datasets} Tox21 is a public dataset \cite{wu2018moleculenet} consisting of 8014 toxicity measurements on 7 nuclear receptor signals and 5 stress response panels \cite{wu2018moleculenet}. Each of the 12 targets might be equal to "1" or "0", representing the "active" or "not active" state. Then, the task for this dataset is multi-task binary classification. 
\begin{figure}[b]
\centering
\includegraphics[height=48mm]{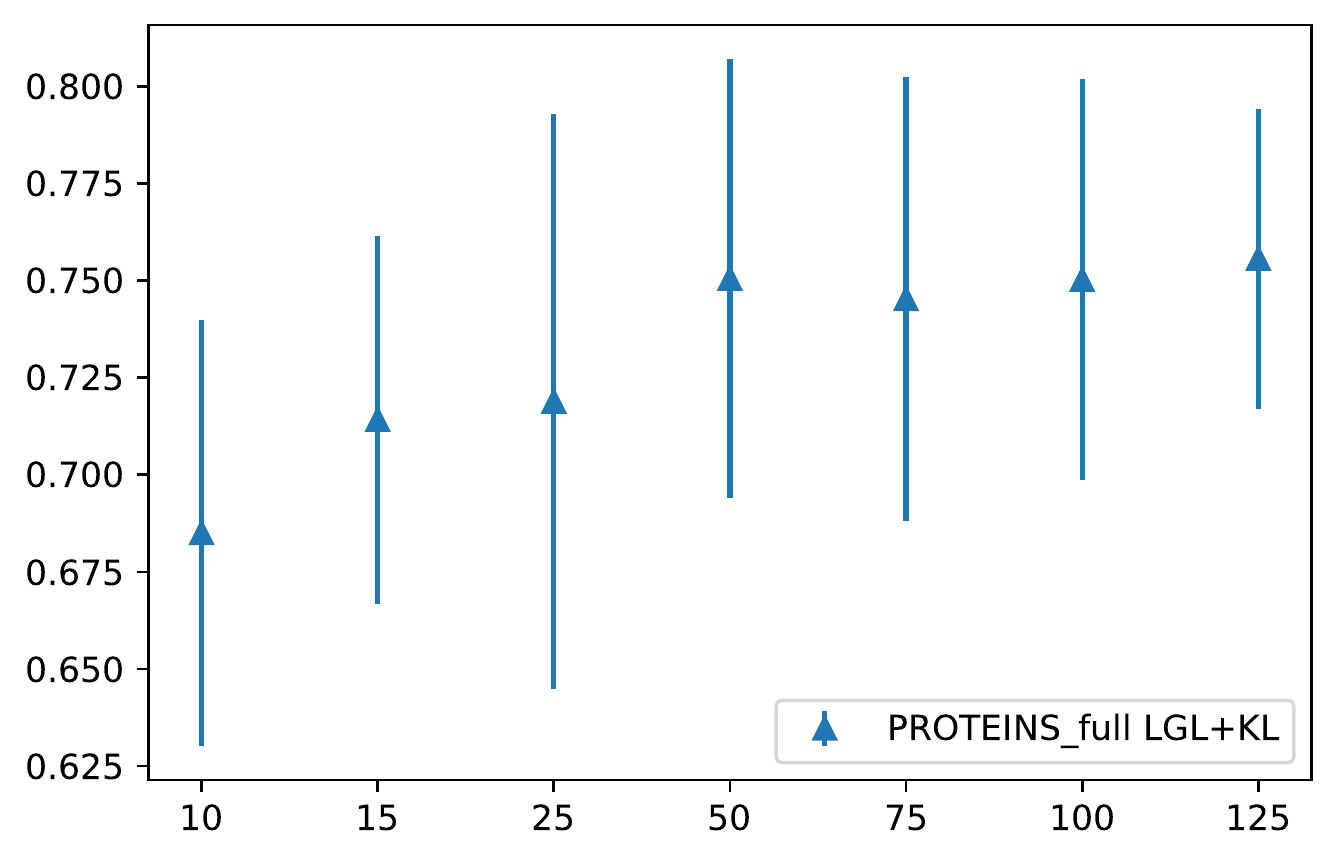}
\caption{PROTEINS$_{3}$. The x-axis shows batch size, y-axis indicates accuracy. Triangles indicate mean accuracy and tails standard deviation.}
\label{fig:proteins_batch_size_comparison}
\end{figure}
\subsection{Implementation details}\label{chapter:implementation_details}
All the experiments are trained with the Adam optimization \cite{kingma2017adam}.
For each dataset number of layers, learning rate, batch size, alpha in \eqref{eq:loss_sum}, and the global pooling operator (mean or add) are tuned based on the validation loss.
In Table \ref{tab:acc_tab}, applications are trained with GraphConv \cite{morris2020weisfeiler} layers in the node-level module. In Table \ref{tab:sota_acc_tab}, we used GIN \cite{xu2019powerful} to show the flexibility of GiG models and its possibility to adapt to the input data.
For classification losses, optimizer, node-level layers and pooling operators, we used the versions implemented in PyTorch \cite{NEURIPS2019_bdbca288} and PyTorch Geometric \cite{pytorch_geometric}.

\textbf{Dataset split.\hspace{1mm}}
The HCP dataset is split into training (72\%), validation (8\%), and test (20\%) sets \cite{PERVAIZ2020116604}. Results are averaged over five consecutive runs of the best model. For PROTEINS$_{29}$, we followed \cite{zhang2019hierarchical} and randomly split the dataset for train and test sets with 90\%, 10\%  and repeat it 10 times. For each run the test set is fixed, then the training set is split to train and validation sets by the k-folds split, where k=10. The final evaluation was based on 10 runs. For the Tox21 dataset, we use predefined scaffold splits from \cite{hu2021open}. The same sets were used for all suggested GiG models and baselines. For fair comparison,  D\&D, NCI1, ENZYMES, and PROTEINS$_3$ we use the same test protocol as in \cite{errica2020fair}.

\textbf{Importance of batch size.\hspace{1mm}}
Our method does not require the full population to be known a-priori, but rather builds the population graph on a subset of samples according to the extracted batch. This has the advantage of both reducing the computational complexity and allowing our method to be used in the more challenging inductive setting, where part of the population is unknown at training time.
As a consequence, the number of input graphs that we consider in a batch has a direct effect on the performance of our method. 
To investigate this aspect, we fixed the batch size to a specific value and optimized the remaining hyperparameters according the validation loss. 
We report in Fig. \ref{fig:proteins_batch_size_comparison} results obtained on PROTEINS$_3$ dataset using our LGL+KL model. Small triangles indicate mean accuracy and tails show the standard deviation. From the plot, we can see that considering more graphs in the population has a benefit up to around 50 samples, after which the performance stabilizes.

\subsection{Quantitative results}
In this subsection, we show experiments on three main and four benchmark datasets. 
To show the benefits of adding the population level module we compared our models with a regular graph convolutional baseline (GCN in Table \ref{tab:acc_tab}) and the state-of-the-art method for each application: HGP-SL \cite{zhang2019hierarchical} for proteins, ElasticNet \cite{ElasticNet} for HCP, and ToxicBlend \cite{zaslavskiy2018toxicblend} for Tox21. 
Further, we investigate three different baseline methods of building the population graph. First, to show the benefits of learning the graph, we suggest building the population level graph sampling a random graph using Erdős–Rényi model with p=10 (Random). 
The second strategy consists on building a KNN graph based on the input graph similarities computed with the WL graph kernel \cite{Weisfeiler_Lehman} with k=10 (KNN). Lastly, as an hybrid approach, we adopt a strategy similar to DGCNN \cite{wang2019dynamic} and dynamically build a KNN graph on the output features of the first module $F_1$ (GiG DGCNN). Despite changing during the training process, with this approach the graph structure is not directly learned as in our method.

\textbf{HCP.\hspace{1mm}}\label{chapter:qualitative_results_hcp}
Table \ref{tab:acc_tab} row 1 shows the results for experiments on gender prediction task on HCP dataset. We compare our results with the state of the art ElasticNet model (non-graph-based)  \cite{PERVAIZ2020116604}. 
The proposed models LGL, LGL+KL outperform the SOTA by 3.92\% and 4.17 \% respectively. 
Further, comparing GCN with population-graph-based models (DGCNN, LGL, LGL+KL) we see that a proper population graph helps with the downstream task and boosts the performance of at least 4.78 \%. In contrast, a comparison of fixed population graphs with GCN shows that the wrong population graph might significantly decrease the final classification performance.

\begin{figure*}[t]
\centering
\begin{subfigure}[b]{.245\textwidth}
  \centering
  \includegraphics[width=0.80\linewidth]{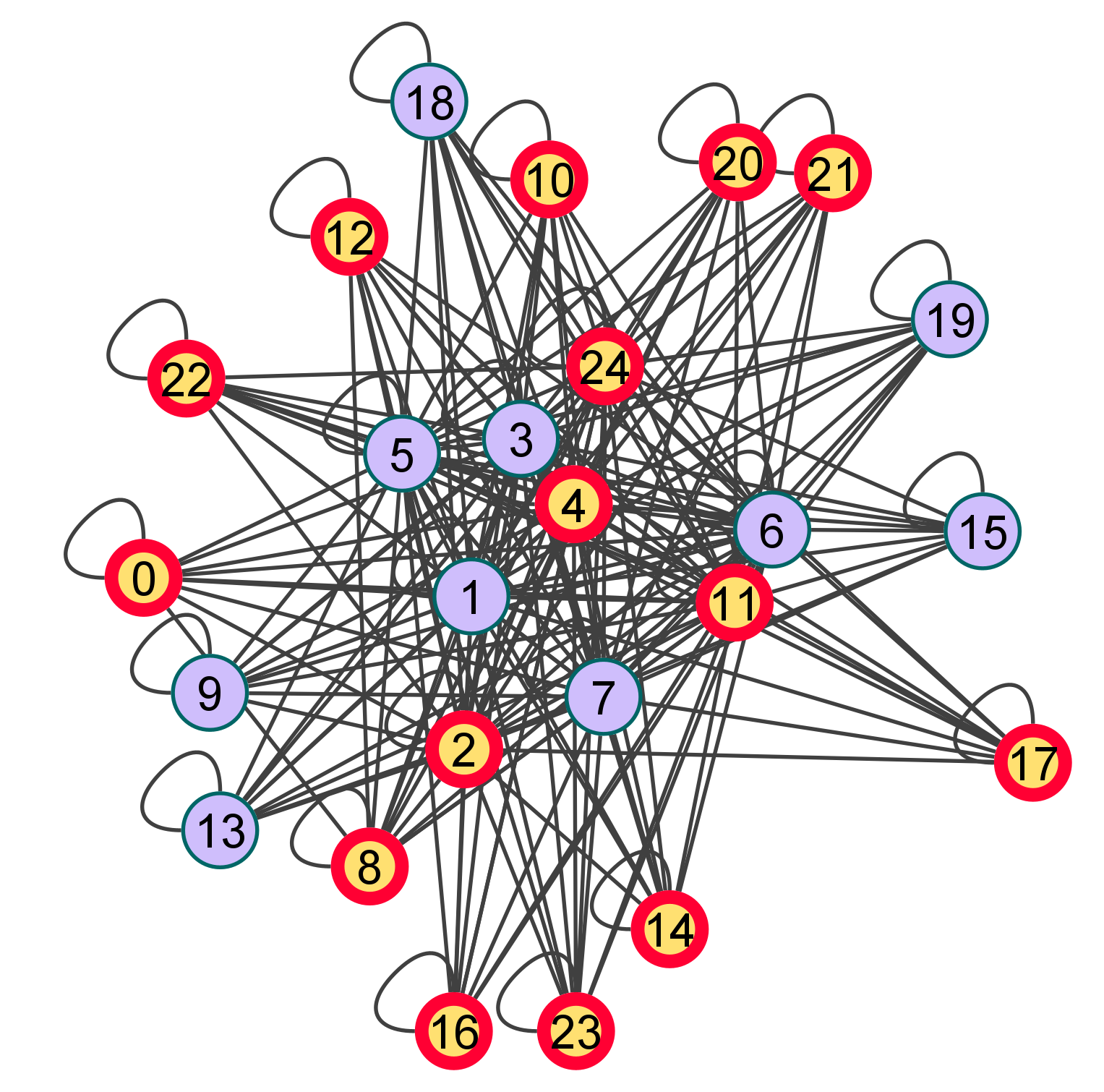}
  \caption{KNN}
  \label{fig:hcp_names_population_graph_comparison_knn}
\end{subfigure}
\begin{subfigure}[b]{.245\textwidth}
  \centering
  \includegraphics[width=0.70\linewidth]{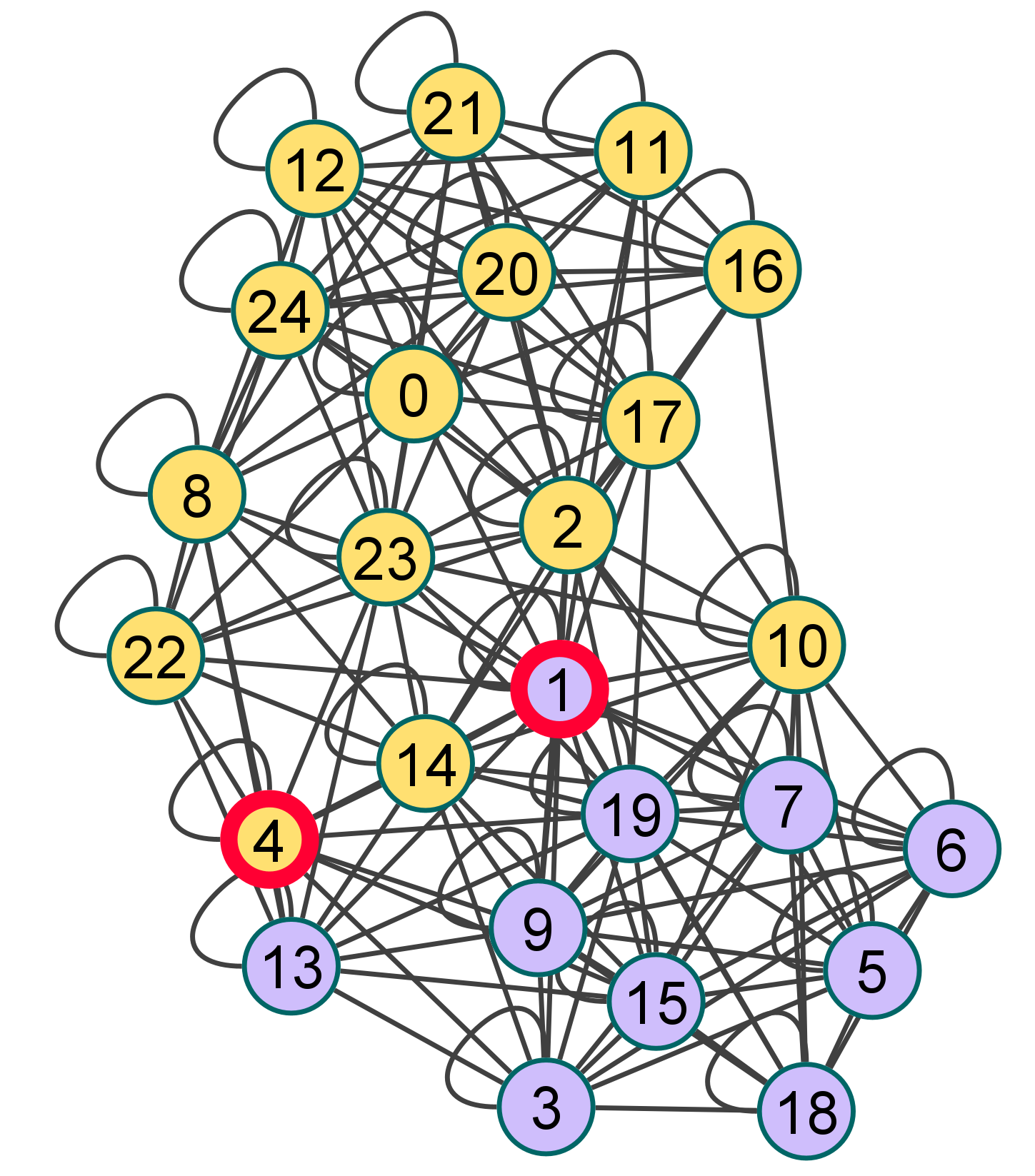}
  \caption{GiG DGCNN}
  \label{fig:hcp_names_population_graph_comparison_dgcnn}
\end{subfigure}
\begin{subfigure}[b]{.245\textwidth}
  \centering
  \includegraphics[width=0.60\linewidth]{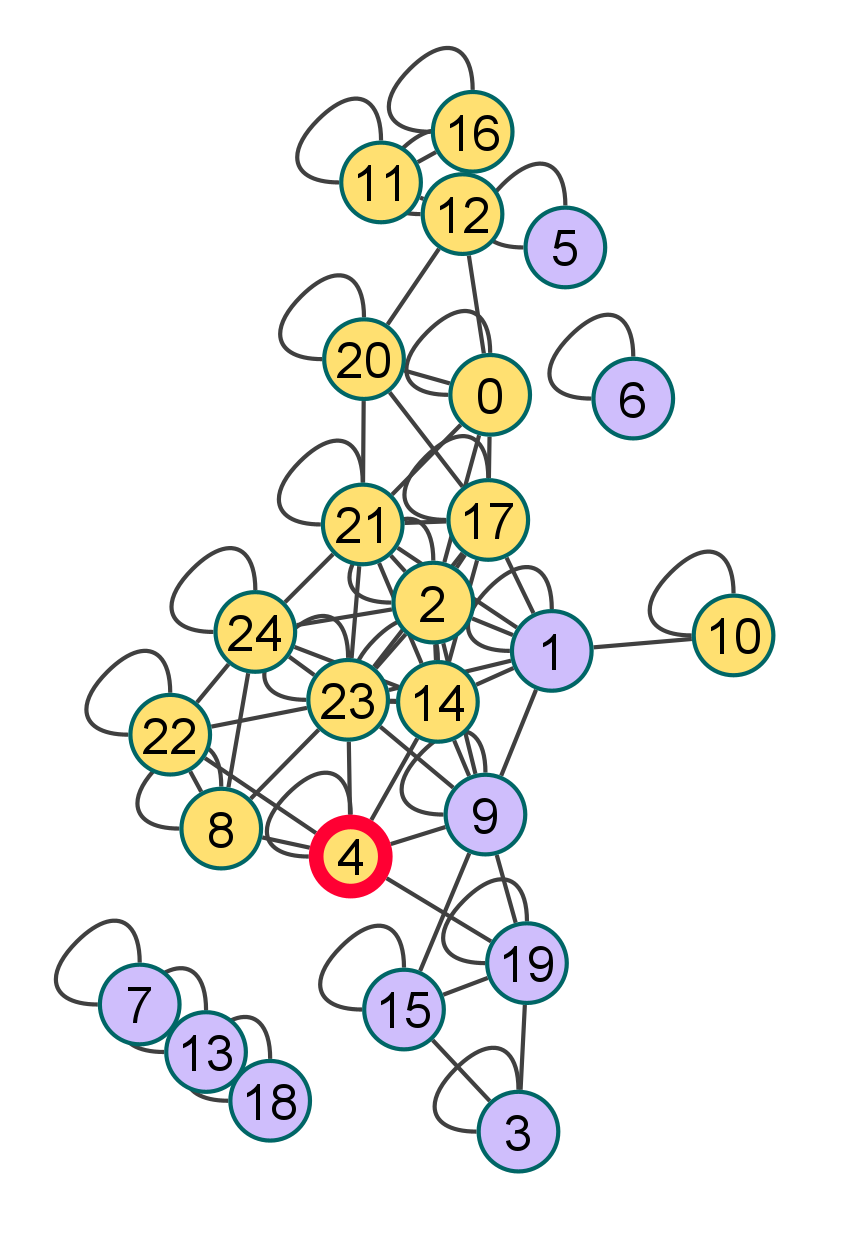}
  \caption{LGL}
  \label{fig:hcp_names_population_graph_comparison_cdgm}
\end{subfigure}
\begin{subfigure}[b]{.245\textwidth}
  \centering
  \includegraphics[width=0.40\linewidth]{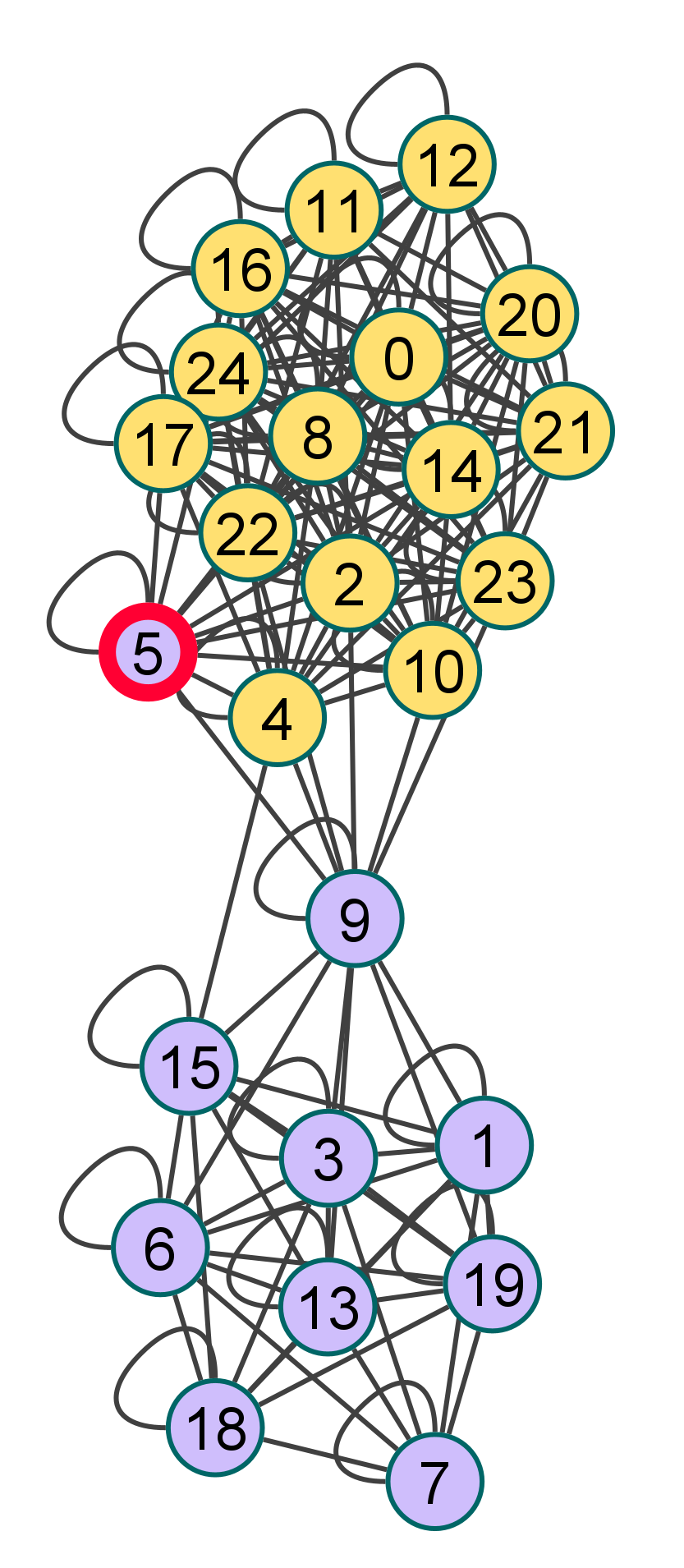}
  \caption{LGL+KL}
  \label{fig:hcp_names_population_graph_comparison_cdgm_kl}
\end{subfigure}
\caption[HCP: population graphs comparison according to misclassified identifiers]{HCP: population graphs comparison according to misclassified identifiers. The location of the node depends purely on the binary structure of the graph. Misclassified nodes are circled in red.}
\label{fig:hcp_names_population_graph_comparison}
\end{figure*}
%
\begin{figure*}[t]
\centering
\begin{subfigure}{.45\textwidth}
  \centering
  \includegraphics[height=40mm]{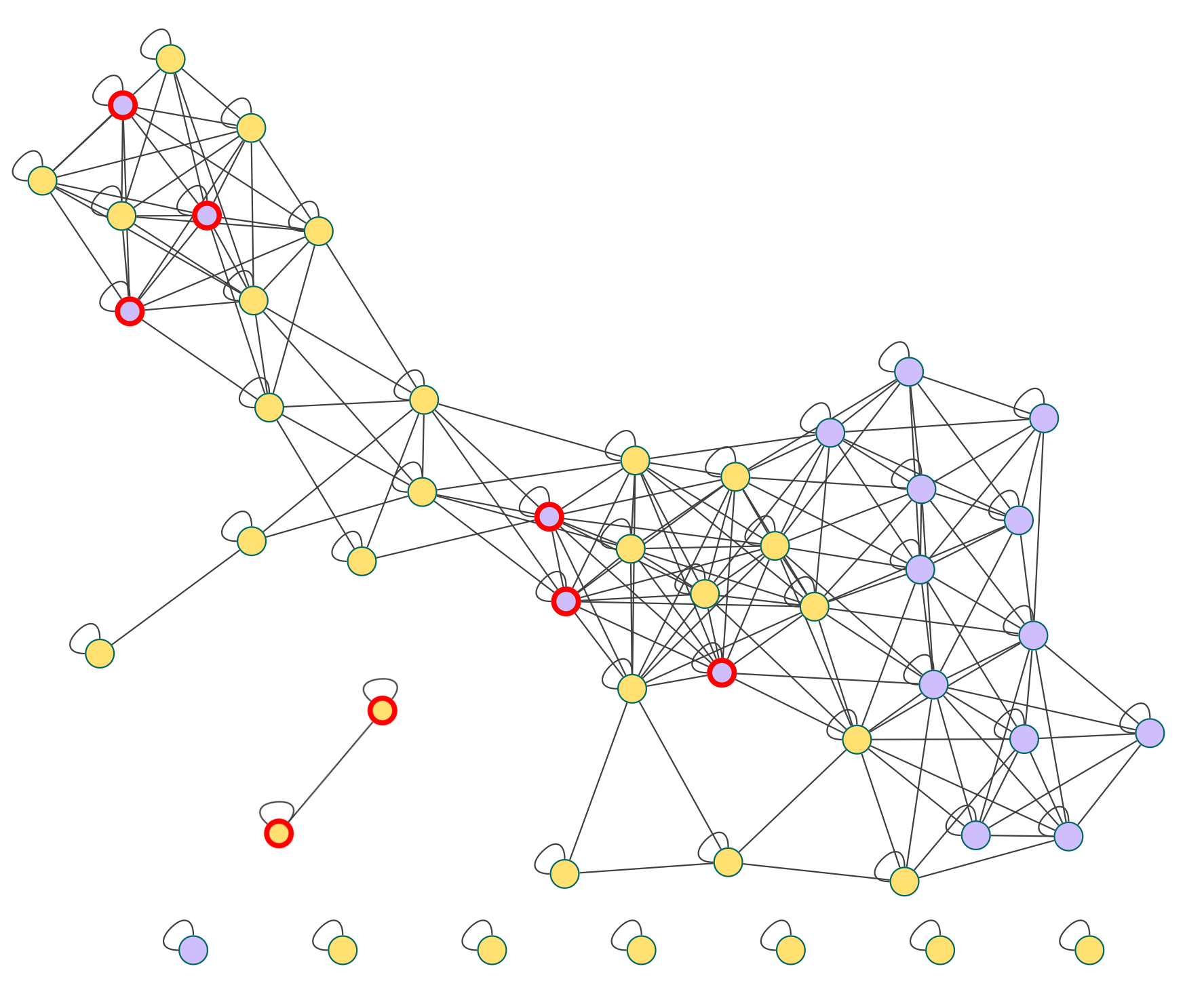}
  \caption{LGL}
  \label{fig:proteins_population_graph_comparison_cdgm}
\end{subfigure}
\begin{subfigure}{.45\textwidth}
  \centering
  \includegraphics[height=40mm]{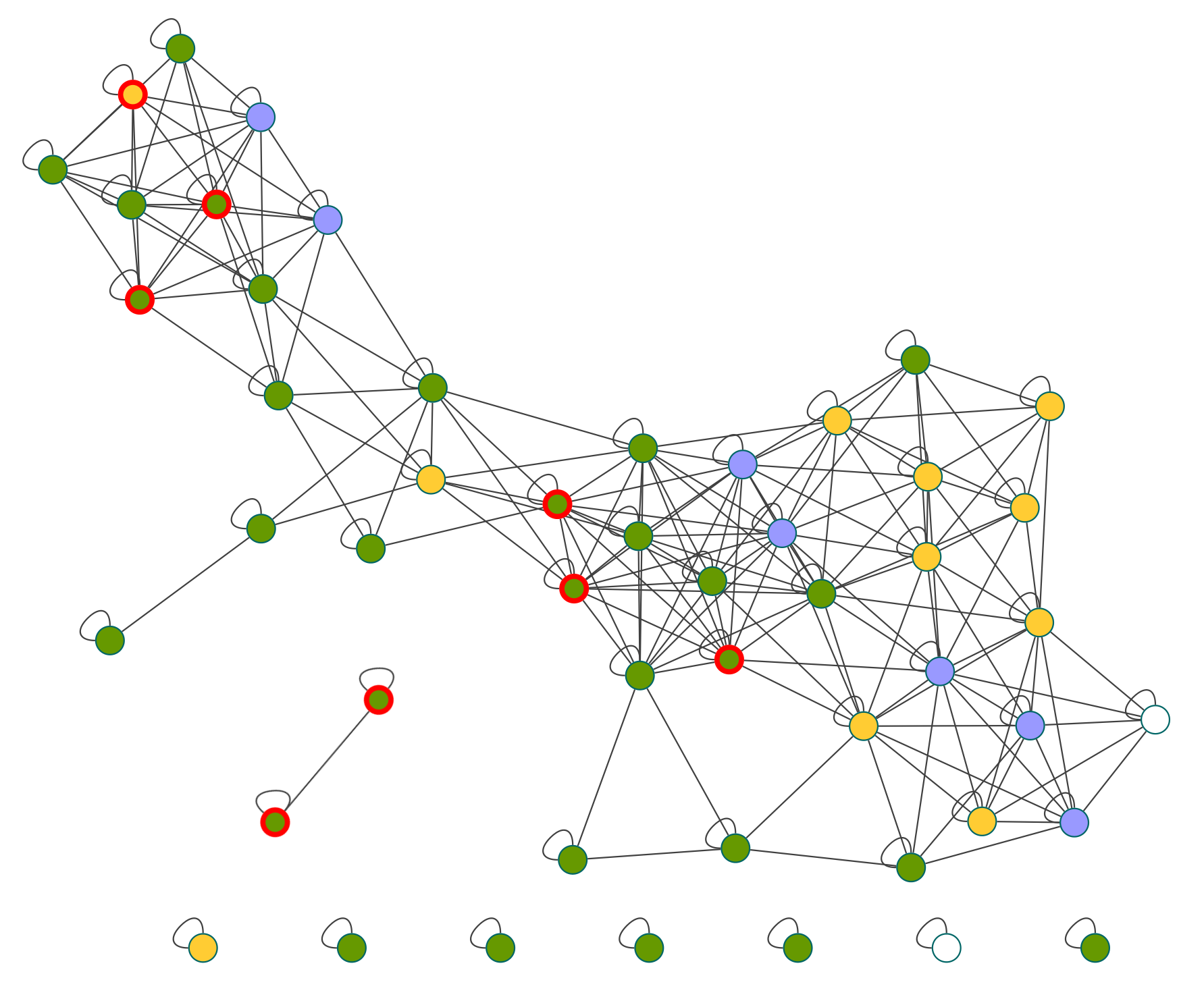}
  \caption{LGL}
  \label{fig:proteins_population_graph_comparison_y1_cdgm}
\end{subfigure}
\begin{subfigure}{.45\textwidth}
  \centering
  \includegraphics[height=17mm]{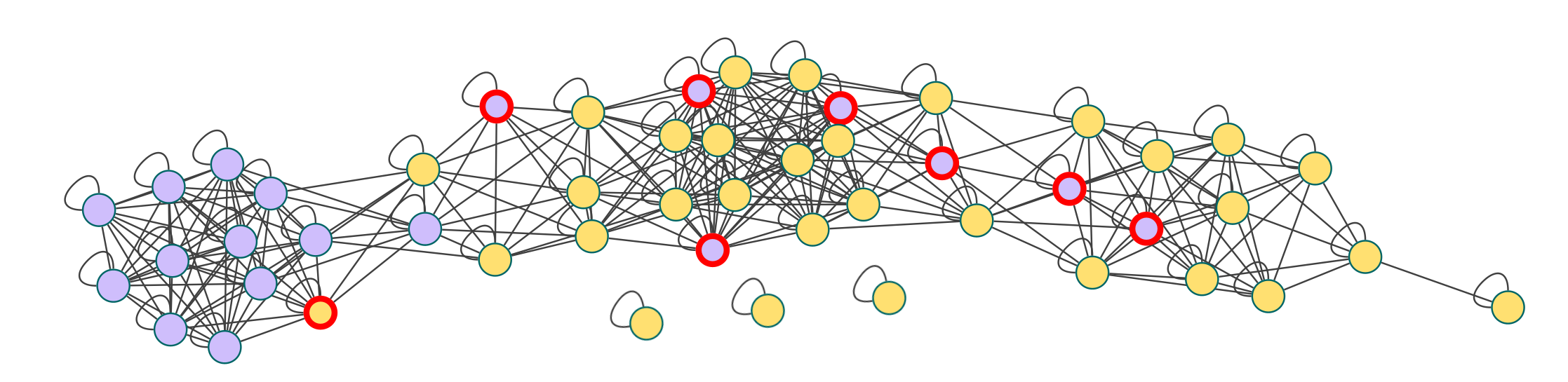}
  \caption{LGL+KL}
  \label{fig:proteins_population_graph_comparison_cdgm_kl}
\end{subfigure}
\begin{subfigure}{.45\textwidth}
  \centering
  \includegraphics[height=17mm]{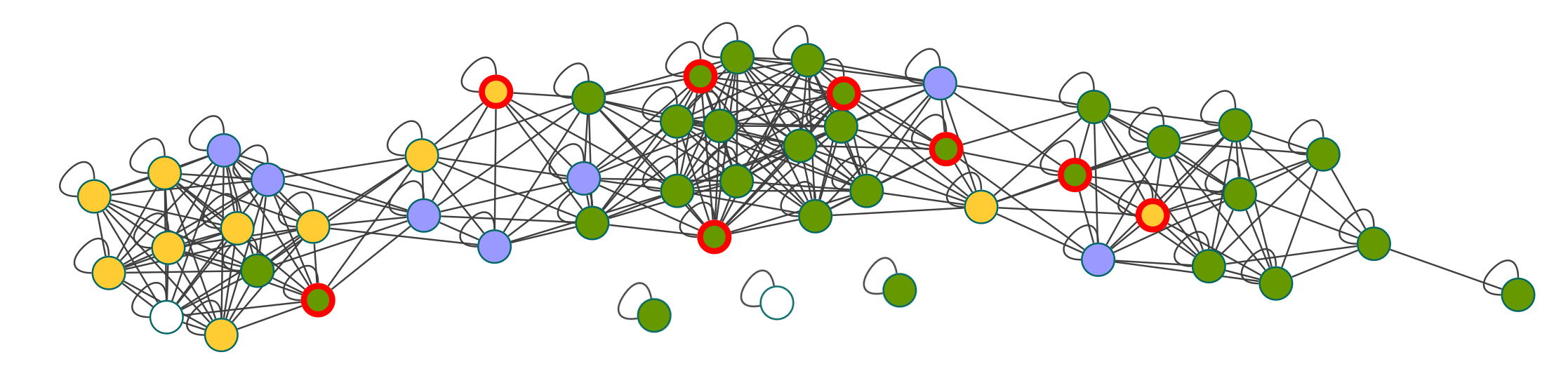}
  \caption{LGL+KL}
  \label{fig:proteins_population_graph_comparison_y1_cdgm_kl}
\end{subfigure}
\caption[PROTEINS$_{29}$: population graphs comparison.]{PROTEINS$_{29}$: population graphs comparison. Misclassified nodes circled by red. The plotting threshold for LGL is equal to 0.01 and for LGL+KL is 0.5 respectively, which means that only edges with weight values bigger than the chosen threshold were plotted. In the first column, yellow represents enzymes (class 0) and violet non-enzymes (class 1). In the second column: yellow, violet and green represent mainly alpha, mainly beta and alpha-beta categories respectively.}
\label{fig:proteins_population_graph_comparison}
\end{figure*}
\begin{table}[t]
\caption[Comparison of results for SOTA datasets]{Comparison with general purpose graph classification methods. Top three  performance are in bold face, blue and red.}
\setlength{\tabcolsep}{4pt}
\resizebox{0.49\textwidth}{!}{\begin{tabular}{@{}l|cccc@{}}\hline
          & D\&D & NCI1 & PROTEINS$_3$ & ENZYMES \\ \hline
Deep-Graph-CNN &     \textcolor{blue}{76.6 ± 4.3} & 76.4 ± 1.7 &  72.9 ± 3.5 & 38.9 ± 5.7 \\ 
DiffPool &  75.0 ± 3.5& 76.9 ± 1.9 &\textcolor{red}{73.7 ± 3.5} & \textcolor{blue}{59.5 ± 5.6}   \\ 
ECC  &      72.6 ± 4.1 & 76.2 ± 1.4 & 72.3 ± 3.4 & 29.5 ± 8.2       \\ 
GIN  &      75.3 ± 2.9 & \textcolor{red}{80.0 ± 1.4} & 73.3 ± 4.0 & \textbf{59.6 ± 4.5}      \\ 
GraphSAGE & 72.9 ± 2.0 & 76.0 ± 1.8 & 73.0 ± 4.5 & \textcolor{red}{58.2 ± 6.0}      \\ 
GiG DGCNN    & 70.4 ± 5.9 & 79.1 ± 2.2 &70.8 ± 4.0 &30.3 ± 8.7     \\ \hline
GiG LGL   & \textcolor{red}{74.9 ± 3.9}   &  \textcolor{blue}{81.0 ± 1.9}    &   \textcolor{blue}{75.0 ± 2.7}       &   54.5 ±   5.6
      \\ 
GiG LGL+KL & \textbf{76.7 ± 4.9}    & \textbf{81.8 ± 1.8}     &    \textbf{75.6 ± 3.9}     &    50.0 ± 7.0     \\ \hline
\end{tabular}}
\label{tab:sota_acc_tab}
\end{table}
\textbf{PROTEINS$_{29}$.\hspace{1mm}}
In the PROTEINS$_{29}$ dataset (Table \ref{tab:acc_tab} row 2), the LGL+KL model performs very close to the SOTA model HGP-SL \cite{zhang2019hierarchical}. The difference in accuracy is 0.13 \%-points, however, LGL+KL is more stable among the  folds and the final standard error is 50\% lower in comparison with HGP-SL. LGL and LGL+KL exceed the GCN by 2 and 4.8 \%-points. This indicates that among graph-based approaches, learning the population-level graph can help with the downstream task. Among all suggested models, LGL+KL gives the best results (84.8 ± 1.08 ), which is higher than LGL by 2.8 \%-points. Regularizing node degrees with the KL-loss stabilizes the learning process and boosts the classification performance.

\textbf{Tox21.\hspace{1mm}}\label{chapter:qualitative_results_tox21}
For the Tox21 dataset (Table \ref{tab:acc_tab} row 3), Random and KNN baselines perform similarly and are close to the SOTA results on scaffold split \cite{zaslavskiy2018toxicblend}. LGL+KL and LGL outperform ToxicBlend \cite{zaslavskiy2018toxicblend} with a gap of 4.97 and 7.5 \%-points. The performance difference between LGL and DGCNN is not significant (0.26 \%). 

\textbf{Bench-marking datasets.\hspace{1mm}} 
The GiG LGL+KL model outperforms the SOTA classification models for D\&D, NCI1 and PROTEINS$_{3}$ dataset, with a margin of 0.1, 1.8, 1.9 \% respectively (Table \ref{tab:sota_acc_tab}). For NCI1 and PROTEINS$_{3}$ datasets LGL is the second best performing model. 
%
%

\begin{figure*}[t]
\centering
\begin{subfigure}{.45\textwidth}
  \centering
  \includegraphics[height=20mm]{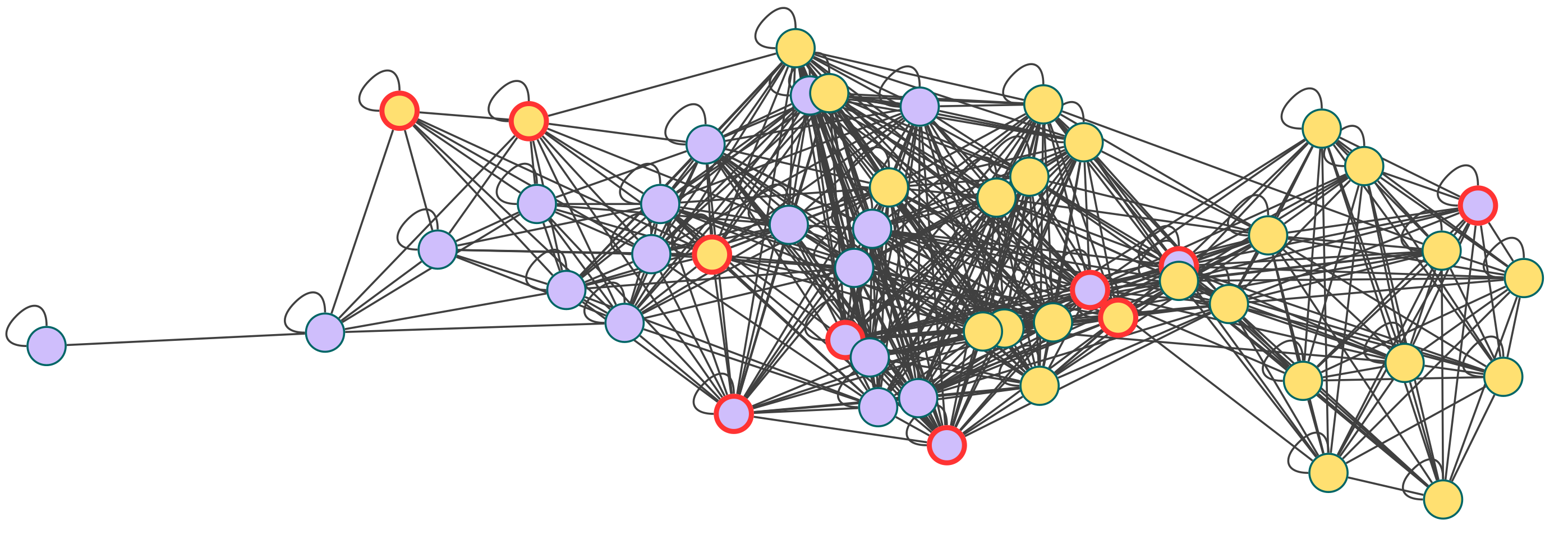}
  \caption{PROTEINS$_{3}$}
  \label{fig:proteins_kl_becnhmark}
\end{subfigure}
\begin{subfigure}{.45\textwidth}
  \centering
  \includegraphics[height=20mm]{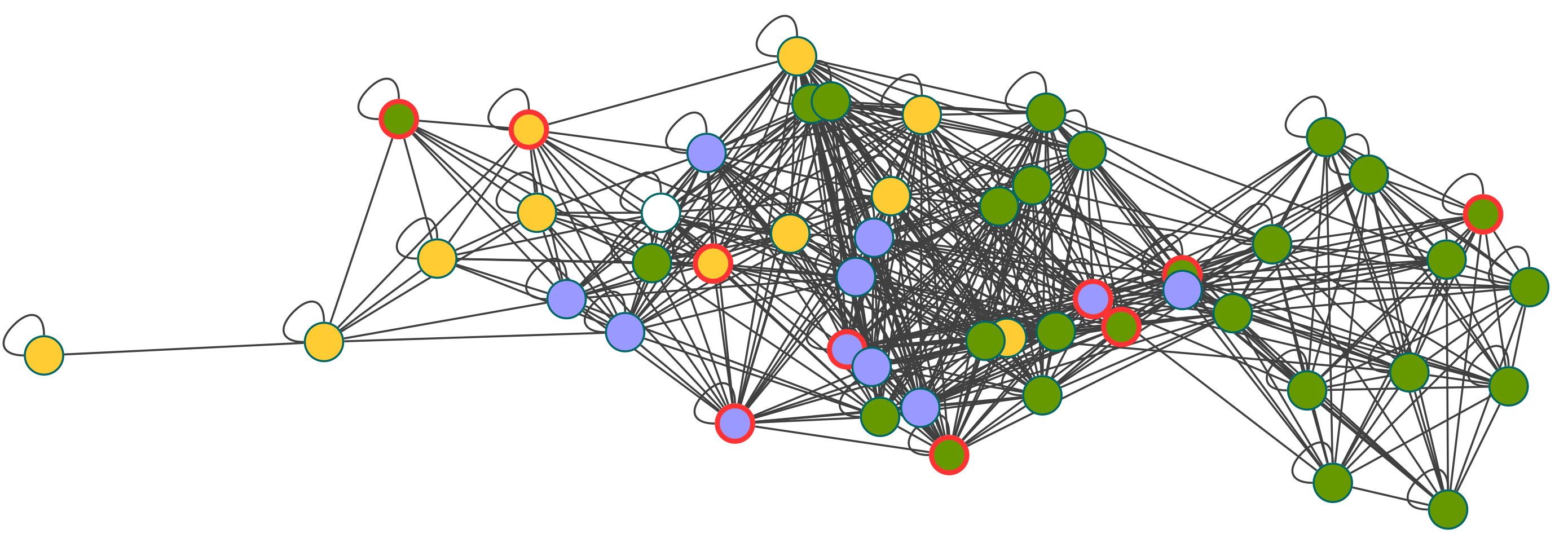}
  \caption{CATH classes, PROTEINS$_{3}$}
  \label{fig:proteins_kl_becnhmark_3}
\end{subfigure}
\begin{subfigure}{.45\textwidth}
  \centering
  \includegraphics[height=22mm]{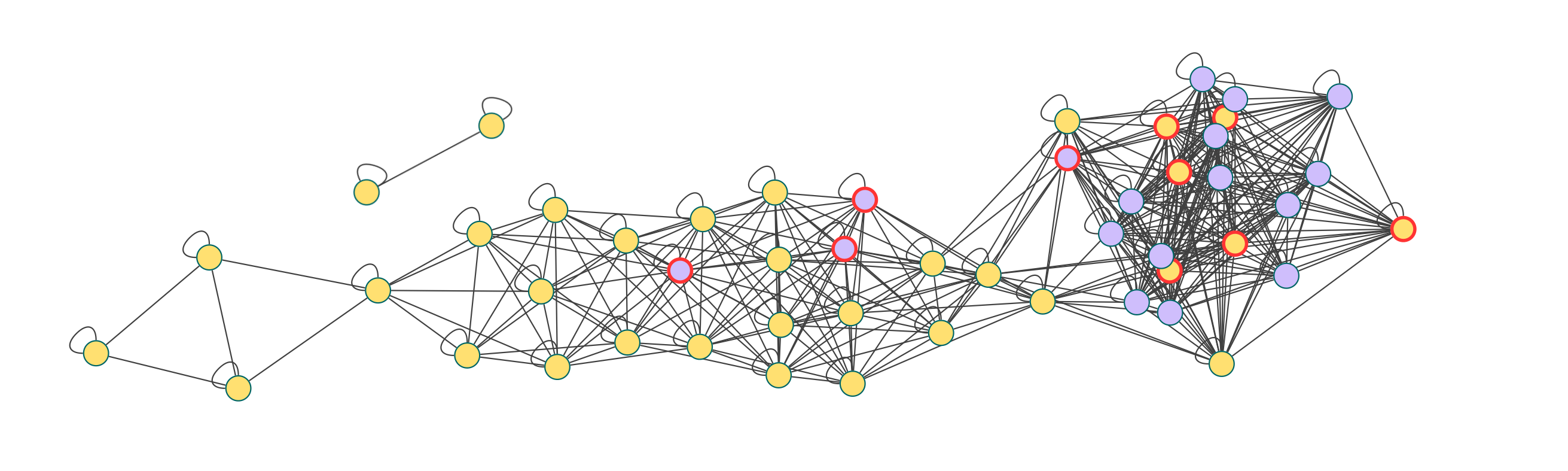}
  \caption{D\&D}
  \label{fig:dd_kl_becnhmark}
\end{subfigure}
\begin{subfigure}{.45\textwidth}
  \centering
  \includegraphics[height=22mm]{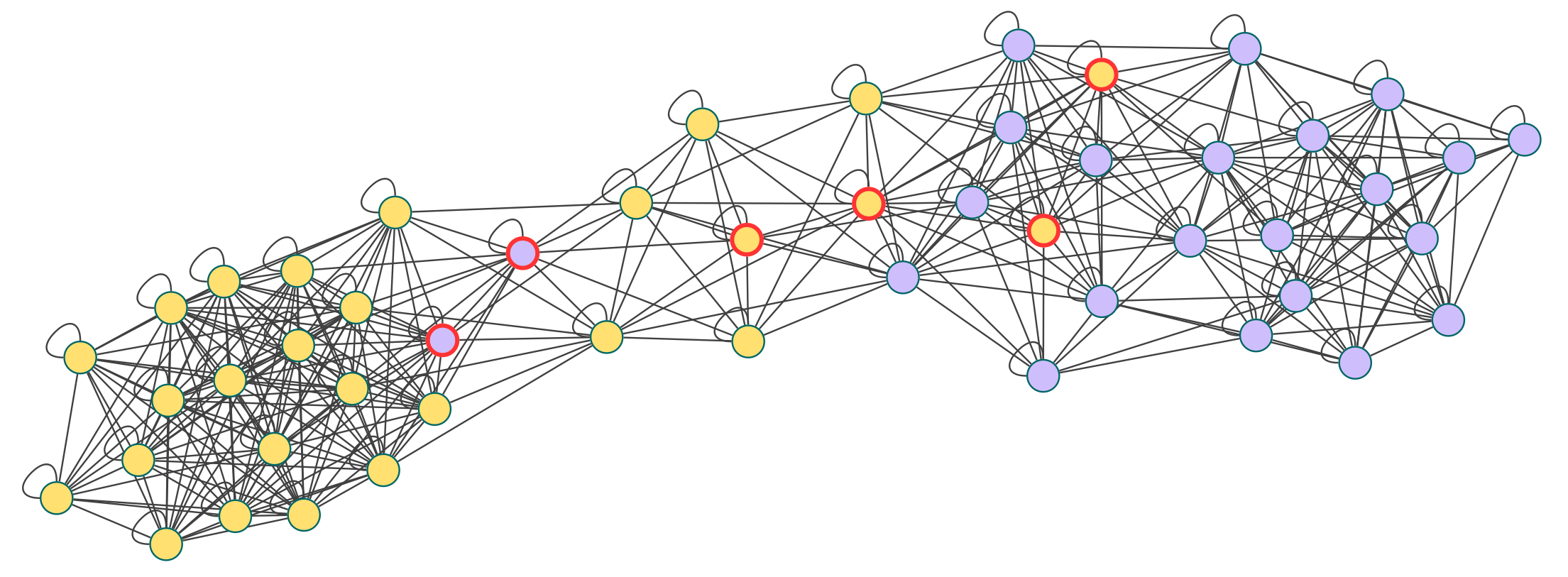}
  \caption{NCI1}
  \label{fig:nci1_kl_becnhmark}
\end{subfigure}
\caption[]{Population graphs comparison of GiG LGL+KL model. Misclassified nodes circled in red. The plotting threshold for LGL+KL is 0.5, which means that only edges with weight values bigger than the chosen threshold were plotted. For two color plots: yellow represents class 0 and violet class 1. CATH hierarchy: yellow, violet and green corresponds to mainly alpha, mainly beta and alpha-beta respectively.}
\label{fig:benchmark_population_graph_comparison}
\end{figure*}
\subsection{Knowledge discovery analysis}
In this section, we show and discuss the population-level graph and how they might be exploited as a source of additional information for data exploration.

\begin{figure}[b!]
\centering
\begin{subfigure}{.5\textwidth}
  \centering
  \includegraphics[height=25mm]{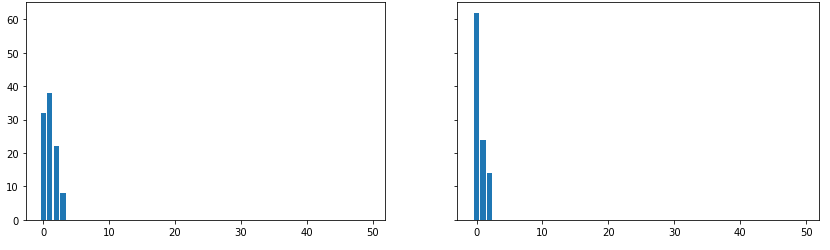}
  \caption{LGL}
  \label{fig:proteins_cdgm_nodes_degree_dist_1}
\end{subfigure}
\begin{subfigure}{.5\textwidth}
  \centering
  \includegraphics[height=28mm]{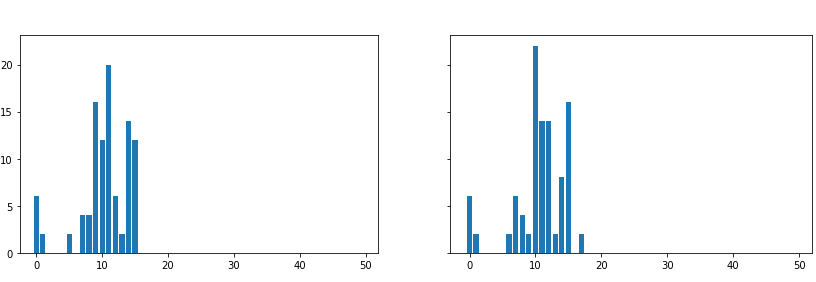}
  \caption{LGL+KL}
  \label{fig:proteins_cdgm_nodes_degree_distr_2}
\end{subfigure}
\caption{PROTEINS$_{29}$ GiG LGL vs GiG LGL+KL nodes degree distribution comparison. The x-axis shows node degrees divided into 50 bins, the y-axis shows the occurrence of nodes degree. Left column indicates normal nodes degree distribution comparison, right column - with thresholds equal to 0.5.}
\label{fig:proteins_nodes_degree_distr}
\end{figure}

\textbf{HCP:}
As can be seen in Fig. \ref{fig:hcp_names_population_graph_comparison_knn}, the KNN baseline (k=10) performs the worst for HCP dataset.  This method misclassifies all nodes with the class "0" and there is no evidence of clustering, all objects are connected and grouped. In comparison, all population graphs from the proposed GiG methods are well clustered. Importantly, the LGL adjacency matrix $\mathbf{A_p}$ of the learned population graph entirely consists of near-zero values, indicating that the model is not confident about the learned graph structure. For interpretability analysis, we plotted LGL in Fig. \ref{fig:hcp_names_population_graph_comparison_cdgm} with an edge weight threshold at a level of $1\mathrm{e}{-1}$. In contrast, the KL-loss resulted in edge weights within the LGL+KL population-level graph is well binarized at a level of $0.5$.
Fig. \ref{fig:hcp_names_population_graph_comparison} shows the unique numbers of each node as well. DGCNN locates nodes number "4" and "1" on the border of two classes and misclassifies them. LGL+KL mislocates the node "5", which in the previous population graphs is located inside its cluster. For all methods except KNN, some nodes tend to be located together. For example nodes with identifiers "11", "12" and "16" as well as "23", "2", "14". This is an indicator that these nodes have some similar properties or common characteristics. 

\textbf{PROTEINS$_{29}$:}
In Fig. \ref{fig:proteins_population_graph_comparison_cdgm} and Fig. \ref{fig:proteins_population_graph_comparison_cdgm_kl},
GiG LGL and GiG LGL+KL show a clustering trend. However, similar to the HCP dataset, the LGL model is quite uncertain about the learned graph connection. It can be seen in Fig. \ref{fig:proteins_weights_degree_distr}, where most adjacency matrix values are located near-zero for LGL. In contrast, using the KL Loss the GiG LGL+KL model is more confident regarding its decision (Fig. \ref{fig:proteins_weights_degree_distr_b}). 
The binarized nodes degree distribution is skewed to the left ( Fig. \ref{fig:proteins_nodes_degree_distr}). The nodes degree distribution of the LGL+KL model remains almost the same after the $0.5$ thresholding, which also indicates the confidence of the model. 
In Fig. \ref{fig:proteins_population_graph_comparison_y1_cdgm}, \eqref{fig:proteins_population_graph_comparison_y1_cdgm_kl} we see that nodes are clustered according to CATH classes: "1", "2", "3", i.e. "mainly alpha", "mainly beta" and "alpha-beta" respectively. Remarkably, most nodes that were misclassified according to task labels, were located on the proper CATH class clusters. This indicates that GiG models not only cluster population-level graphs according to training labels but as well capture the structure and properties of the input data. Moreover, the majority of prediction errors were done on the proteins that belong to alpha-beta CATH classes. This is explainable with domain knowledge: "alpha-beta" protein structures are composed of $\alpha$-helices and $\beta$-strands, which are combined inside the structure of "mainly alpha" or "mainly beta" proteins. So it is more challenging for the model to find a proper location inside the population graph for these input objects. 

\textbf{Tox21:}
Tox21 is a multi-task dataset, where assays are highly correlated with each other and training models towards only one of the toxic tests is less effective \cite{doi:10.1021/acs.jcim.0c01294}. However, analyzing the population graph with multiple labels is challenging. Therefore, in this work, we only provide the quantitative results for the Tox21 dataset. From those, it is evident that incorporating the population graph helped in improving the final downstream task performance.

\textbf{Bench-marking datasets:} The population-level graphs from GiG LGL+KL for NCI1, D\&D and Proteins$_3$ are shown in the Fig. \ref{fig:benchmark_population_graph_comparison}. The misclassification of some nodes is quite expected, given their location in $\mathbf{G_p}$ and obtained via GCN based model node representations. Explaining these locations requires some additional information, like CATH hierarchy for PROTEINS dataset. From PROTEINS$_3$ population graphs, it is evident that most misclassified nodes are embedded at locations that correspond to CATH classes. This indicates that population graphs and their node embedding provide a form of knowledge discovery, which corresponds to additional domain information of the input objects.

\section{Discussion}\label{section:discussion}
To learn the population graph in LGL-based models, we utilize a soft-threshold which is tuned by two learnable parameters: $temp$ and $\theta$. One future direction could be to learn the $\theta$ values per node. The latter hugely affects the adjacency matrix values and might help to push them to either 0 or 1. Even though these parameters are learned, the proper initialization is crucial and might lead to different population graphs and different classification performances. The overall sparsity of the graph depends on the value of the threshold $\theta$.  
During our experiments, we also noticed that it is crucial to obtain proper input-graph representations, since the population graph is learned or constructed based on them. Thus the node-level module and pooling operator should be adapted to the application and the input data. Another crucial point is the target distribution with learnable parameters in the GiG LGL+KL model. In this work, we utilized a normal distribution to regularize the latent graph. Thus, directions for future work include the investigation of the influence of different target distributions and their reasoning, as well as considering graph pooling operators. In addition, all constructed and learned latent graphs were obtained based on similarities between input objects, however, one more relevant and challenging application for this work might be to learn the graph that represents Protein-protein interactions (PPI). We would like to point out a few limitation of the proposed work. 1) The choice of the target distribution needs further exploration. In this paper we only considered Gaussian distribution. 2) Even though the model gives us the freedom of choice for $F_1$ and $F_2$, a careful selection of $F_1$ and $F_2$ is necessary. 
\section{Conclusion}\label{section:conclusion}
In this paper, we propose a method for learning a latent graph structure on graph-valued input data. The latent graph provides a form of knowledge discovery on the distribution and neighborhood relationships between input samples. Further, the latent graph is learned end-to-end, along with the downstream task, e.g. classification. More precisely, we proposed a structure consisting of three main modules: node-level module, population-level module, and GCN classifier that might be easily adapted to any graph classification application. Based on our experiments, we noticed that learning the population graph mainly increases (HCP, Tox21, NCI1, D\&D, PROTEINS$_3$ datasets) the performance of the downstream task. Additionally, the latent graph might be used for understanding the made decision (PROTEINS) and discovering new information based on interconnected objects. Moreover, we suggested regularising the latent graph's node degree distribution, in order to obtain a better representation for the graph-valued input population.
\bibliographystyle{IEEEtran}
\bibliography{tmi}

\end{document}